\documentclass[11pt,a4paper]{article}
\usepackage[hyperref]{acl2021}
\usepackage{times}
\usepackage{latexsym}

\usepackage{mathtools}
\usepackage{paralist}
\usepackage{hyperref}
\usepackage{booktabs}
\usepackage{enumitem}
\usepackage{multirow}
\usepackage{makecell}

\usepackage{amssymb}
\usepackage{pifont}
\newcommand{\cmark}{\ding{51}}%
\usepackage{microtype}

\aclfinalcopy 

\title{What can Neural Referential Form Selectors Learn?}

\author{Guanyi Chen\textsuperscript{$\spadesuit$}\thanks{\hspace{0.2cm}Equal contribution}, 
Fahime Same\textsuperscript{$\heartsuit$}\footnotemark[1], \and
Kees van Deemter\textsuperscript{$\spadesuit$}\\
\textsuperscript{$\spadesuit$}Department of Information and Computing Sciences, Utrecht University\\
\textsuperscript{$\heartsuit$}Department of Linguistics, University of Cologne\\
\texttt{g.chen@uu.nl, f.same@uni-koeln.de, c.j.vandeemter@uu.nl}}

\date{}

\begin{document}
\maketitle
\begin{abstract}
Despite achieving encouraging results, neural Referring Expression Generation models are often thought to lack transparency. We probed neural Referential Form Selection (RFS) models to find out to what extent the linguistic features influencing the RE form are learnt and captured by state-of-the-art RFS models. The results of 8 probing tasks show that all the defined features were learnt to some extent. The probing tasks pertaining to referential status and syntactic position exhibited the highest performance. The lowest performance was achieved by the probing models designed to predict discourse structure properties beyond the sentence level. 

\end{abstract}

\section{Introduction}\label{sec:intro}

Referring Expression Generation (REG) is one of the main stages of classic Natural Language Generation (NLG) pipeline~\citep{reiter_dale_2000, krahmer-van-deemter-2012-computational, van2016computational}. REG studies are concerned with two different tasks. The goal of the classic REG task (also called one-shot REG), is to find a set of attributes to single out a referent from a set of competing referents. The second REG task (henceforth discourse REG) is concerned with the generation of referring expressions (RE) in discourse context. \citet{belz-varges-2007-generation} phrase it as follows:
\emph{Given an intended referent and a discourse context, how do we generate appropriate referential expressions (REs) to refer to the referent at different points in the discourse?}

Classic discourse REG was usually understood as a two-step procedure. In the first step, the referential form (RF, i.e, the syntactic type) is determined. For instance, when referring to Joe Biden at a given point in a discourse, the first step is to decide whether to use a proper name (``\emph{Joe Biden}"), a description (``\emph{the president of the USA}"), a demonstrative (``\emph{this person}") or a pronoun (``\emph{he}"). The second step is to determine the RE content, that is, to choose between all the different ways in which a given form can be realised. For instance, to generate a description of Joe Biden, one needs to decide whether to only mention his job (e.g., \emph{The president} entered the Oval Office.), or to mention the country as well (e.g., \emph{The president of the United states} arrived in Cornwall for the G7 Summit.) 

In earlier works, computational linguists linked REG to linguistic theories and built discourse REG systems on the basis of linguistic features.
For example, \citet{henschel2000pronominalization} investigated the impact of 3 linguistic features namely 
recency, subjecthood, and discourse status on pronominalization, i.e. deciding whether the RE should be realised as a pronoun. 
Using these features, they used the notion of \emph{local focus} as a criterion for detecting the set of referents that can be pronominalised.
The same holds for feature-based models (see \citet{belz2010generating} for an overview) where models are trained on linguistically encoded data. 

\begin{table*}[t!]
\small
\centering
\begin{tabular}{p{15cm}}
\toprule
\textbf{Triples}: (AWH\_Engineering\_College, country, India), 
(Kerala, leaderName, Kochi),
(AWH\_Engineering\_College, academicStaffSize, 250),
(AWH\_Engineering\_College, state, Kerala),
(AWH\_Engineering\_College, city, ``Kuttikkattoor''),
(India, river, Ganges) \\
\midrule
\textbf{Text}: AWH Engineering College is in Kuttikkattoor, India in the state of Kerala. The school has 250 employees and Kerala is ruled by Kochi. The Ganges River is also found in India. \\
\textbf{Delexicialised Text}: \textcolor{blue}{\underline{AWH\_Engineering\_College} is in \underline{``Kuttikkattoor''} , \underline{India} in the state of \underline{Kerala} . }\textbf{\underline{AWH\_Engineering\_College}} \textcolor{green}{has 250 employees and \underline{Kerala} is ruled by \underline{Kochi} . The \underline{Ganges} River is also found in \underline{India} .}\\
\bottomrule
\end{tabular}
\caption{An example data from the \textsc{w}eb\textsc{nlg} corpus. In the delexicalised text, every entity is \underline{underlined}, the target entity is \textbf{boldfaced}, the pre-context is coloured in \textcolor{blue}{blue}, and the pos-context is coloured in \textcolor{green}{green}.}
\label{tab:sample}
\end{table*}

More recently, a number of neural network-based REG models have been presented~\citep{castro-ferreira-etal-2018-neuralreg, cao-cheung-2019-referring, cunha-etal-2020-referring}, where they propose to generate REs in an End2End manner without any feature engineering. 
They all used a benchmark dataset called \textsc{w}eb\textsc{nlg}.
These models generally follow the sequence-to-sequence framework~\citep{NIPS2014_a14ac55a}, where there is an encoder for encoding the given discourse, and a decoder responsible for generating REs using the encoded information. 
The evaluation results suggested that these neural methods perform well not only for selecting the proper RFs, but also for producing fluent REs.
However, it was unclear to what extent these neural models 
can encode linguistic features.

To conduct model inspection, we introduce a series of probing tasks. 
Using probing tasks is a well-established method to analyse whether a model's latent representation encodes specific information. This approach has been widely used for analysing models in machine translation \citep{belinkov-etal-2017-neural}, 
language modelling \citep{giulianelli-etal-2018-hood}, relation extraction \citep{alt-etal-2020-probing}, and so on. Additionally, there had been various works on coreference resolution and bridging anaphora \citep{sorodoc-etal-2020-probing,pandit-hou-2021-probing} which, similar to this paper, target the understanding of reference. More precisely, for a probing task, a diagnostic classifier is trained on representations from the model. 
Its performance embodies how well those representations encode the information associated with the probing task. The aim of this paper is to understand what linguistic features neural models encode when modelling REs.

Our main focus is on the encoding of linguistic features in the representations. In the linguistic tradition, the majority of RE production studies focus on Referential Form Selection (RFS), rather than RE content realisation. 
Our focus in the present work is likewise on RFS.
To tackle RFS, we adopt the state-the-of-the-art neural REG model of~\citep{castro-ferreira-etal-2018-neuralreg}. Additionally, to make comparison, we also propose (1) a strong baseline that uses only a single encoder (while~\citet{castro-ferreira-etal-2018-neuralreg} used multiple encoders); and (2) to leverage pre-trained word embeddings (e.g., GloVe) or language models (e.g., BERT).

Therefore, in this paper, we first introduce the task of RFS on the basis of \textsc{w}eb\textsc{nlg} corpus, and propose a number of neural models to tackle the task.
Subsequently, we introduce 8 probing tasks, each of which is associated with a linguistic feature influencing the choice of RF.
We examine our RFS models on these probing tasks in order to interpret and explain their behaviour. 
The code of each RFS model and the probing classifier is available at: \url{github.com/a-quei/probe-neuralreg}.





\section{Background}\label{sec:background}


\subsection{Discourse REG}

Given a text whose REs have not yet been generated, and given the intended referent for each of these REs, the discourse REG
task is to build an algorithm that generates all these REs.
So far, this task has attracted many research efforts (e.g., \citet{hendrickx-etal-2008-cnts, greenbacker2009udel}) and it has been used in the GREC shared tasks~\citep{belz2010generating}.

More recently, this task was formulated into a format that goes together well with deep learning:
\citet{castro-ferreira-etal-2018-neuralreg} introduced the End2End REG task, built a corresponding dataset based on \textsc{w}eb\textsc{nlg}~\citep{castro-ferreira-etal-2018-enriching}, and constructed NeuralREG models.

The \textsc{w}eb\textsc{nlg} corpus was originally designed to assess the performance of NLG systems~\citep{gardent-etal-2017-creating}. 
Each sample in this corpus contains a knowledge base described by a Resource Description Framework (RDF) triple (Table~\ref{tab:sample}).
\citet{castro-ferreira-etal-2018-neuralreg} and \citet{castro-ferreira-etal-2018-enriching} enriched and delexicalised the corpus to fit the discourse REG task. Table~\ref{tab:sample} shows a text created from a RDF, and its corresponding delexicalised version. 

Taking the delexicalised text in Table~\ref{tab:sample} as an example, given the entity ``\emph{AWH\_Engineering\_College}'', REG chooses a RE based on that entity and its pre-context (``\emph{AWH\_Engineering\_College is in ``Kuttikkattoor'' , India in the state of Kerala . }'') and its pos-context (``\emph{has 250 employees and Kerala is ruled by Kochi . The Ganges River is also found in India .}'').

\subsection{Factors that Influence RE Production} \label{sec:factor}
Languages display a large inventory of expressions for referring to entities \citep{von2019discourse}. In linguistics, the realisation choice a speaker makes has been associated with the accessibility, i.e. activation of mental
representations of a referent at a particular point in discourse: attenuated forms such as pronouns are often used to refer to highly accessible or highly activated referents, while richer forms such as descriptions and proper names are employed in referring to less accessible ones \citep{ariel1990accessing,gundel1993cognitive}.
Due to the central role of referring in communication, a wealth of research has tried to assess the influence of different features modulating the accessibility of a referent. \citet{von2019discourse} refer to these features as \emph{prominence-lending cues}, meaning that they increase the prominence status of their respective referents to some extent. In this section, we merely talk about the ones which will be taken up in our probing experiments, and will not further discuss cues such as animacy \citep{fukumura2011effect}, competition \citep{arnold2010speakers} and coherence relations \citep{kehler2008coherence}.

\emph{Referential status} or \emph{givenness} has been widely discussed in the literature (see \citet{chafe1976givenness,prince1981towards}). When a new character is introduced into the discourse, the chance that this happens by means of a pronoun is slim (unless the referent is situationally given). Pronouns are reserved for referring to previously introduced (or given) referents. 

\emph{Recency}, another well-studied cue, is defined as the distance between the target referent and its antecedent. If a referent is not too far apart from its antecedent, then reduced forms are typically employed to refer to it. 

There are also intra-clausal cues such as \emph{grammatical role} \citep{brennan1995centering} and \emph{thematic role} \citep{arnold2001effect} which impact the prominence status of referents. For instance, the subject of a sentence is perceived to be more prominent than the object.

Discourse-structural features affect the organisational aspects of discourse. Centering-based theories \cite{grosz1995centering} often use the notion of local focus to account for pronominalisation. \emph{Local focus} takes the current and previous utterance into account. \emph{Global focus}, on the other hand, situates a referent in a larger space, namely the whole text or a discourse segment \citep{hinterwimmer2019prominent}. Concepts such as the importance of a referent or familiarity are associated with the global prominence status of entities \citep{siddharthan2011information}. 



\section{Neural Referential Form Selection} \label{sec:model}

In this section, we define the task of RFS built on the \textsc{w}eb\textsc{nlg} dataset, and introduce a number of NeuralRFS models.

\subsection{The RFS Task}


\begin{table}[t!]
\small
\centering
\begin{tabular}{lp{5cm}}
\toprule
Type & Classes \\ \midrule
4-Way & Demonstrative, Description, Proper Name, Pronoun \\
3-Way & Description, Proper Name, Pronoun \\
2-Way & Non-pronominal, Pronominal \\
\bottomrule
\end{tabular}
\caption{3 different types of RF classification.}
\label{tab:cls}
\end{table}
\begin{table*}[t!]
\small
\centering
\begin{tabular}{lccccccccc}
\toprule
 & \multicolumn{3}{c}{4-way} & \multicolumn{3}{c}{3-way} & \multicolumn{3}{c}{2-way} \\\cmidrule(lr){2-4} \cmidrule(lr){5-7} \cmidrule(lr){8-10}
Model & Precision & Recall & F1 & Precision & Recall & F1 & Precision & Recall & F1 \\ \midrule
\texttt{XGBoost} & 53.77 & 51.98 & 51.55 & 71.27  & 69.24  & 68.34  & 86.64  & 82.76  & 84.57  \\ \midrule
\texttt{c-RNN} & \underline{68.79} & \underline{62.95} & \underline{64.96} & \underline{84.49} & \underline{82.52} & \textbf{83.63} & \underline{90.31} & 88.01 & 89.09 \\
\texttt{ +GloVe} & \textbf{69.10} & \textbf{63.90} & \textbf{65.40} & 84.29 & \textbf{82.55} & 83.30 & 89.33 & 88.02 & 88.63 \\
\texttt{ +BERT} & 62.63 & 61.80 & 62.15 & 83.02 & 81.44 & 82.15 & \textbf{90.98} & 88.00 & \textbf{89.42} \\
\texttt{ConATT} & 67.42 & 62.39 & 64.07 & \textbf{85.04} & 82.21 & \underline{83.53} & 89.30 & \textbf{89.19} & \underline{89.23} \\
\texttt{ +GloVe} & 65.98 & 62.49 & 63.67 & 83.62 & 81.41 & 82.45 & 89.60 & \underline{88.06} & 88.80\\
\bottomrule
\end{tabular}
\caption{Evaluation results of our RFS systems on \textsc{webnlg}. Best results are \textbf{boldfaced}, whereas the second best results are \underline{underlined}.}
\label{tab:cls_result}
\end{table*}

Akin to REG, given the previous context $x^{(pre)} = \{w_1, w_2, ..., w_{i-1}\}$ (where $w$ is either a word or a delexicalised entity label), the target referent $w^{(r)} = \{w_i\}$,  and the post context $w^{(pos)} = \{w_i, w_{i+1}, ..., w_n\}$, a RFS algorithm aims at finding the proper RF $\hat{f}$ from a set of $K$ candidate RFs $\mathcal{F} = \{f_k\}_{k=1}^K$.

Regarding the possible RFs for the RFS task, we test 3 different classifications, depicted in  Table \ref{tab:cls}.  Due to the small number of demonstrative noun phrases in the dataset, we decided to also conduct a 3-way classification in which descriptions and demonstratives are merged. Also, most emphasis in the linguistic literature is on the pronominalisation issue. Therefore, we also included a 2-way classification task in the study.

As stated, the main goal of the paper is to understand which linguistic features are encoded by RFS neural models. Additionally, we were curious whether models trained solely for pronominalisation capture different contextual features in comparison with the other two classifications. 

\subsection{NeuralRFS Models}

We build NeuralRFS models by (1)
adopting the best NeuralREG model from \citet{castro-ferreira-etal-2018-neuralreg}, and (2) proposing a new alternative which is simpler, and can easier incorporate pre-trained representations.

\paragraph{\texttt{ConATT}.} We adopt the \texttt{CATT} model from~\citet{castro-ferreira-etal-2018-neuralreg}, which achieves the best performance on REG among the models they tested in their study. Given the inputs, we first use Bidirectional GRU~\citep[BiGRU,][]{cho-etal-2014-learning} to encode $x^{(pre)}$ as well as $x^{(pos)}$. Formally, for each $k\in[pre, pos]$, we encode $x^{(k)}$ to $h^{(k)}$ with a BiGRU: $h^{(k)} = \mbox{BiGRU}(x^{(k)})$. Subsequently, different from~\citet{castro-ferreira-etal-2018-neuralreg}, we encode $h^{(k)}$ into the context representation $c^{(k)}$ using self-attention~\citep{yang-etal-2016-hierarchical}. Concretely, given the total $N$ steps in $h^{(k)}$, we first calculate the attention weight $\alpha^{(k)}_j$ at each step $j$ by:
\begin{equation}
    e_j^{(k)} = v_a^{(k)T}\mbox{tanh}(W_a^{(k)}h_j^{(k)}),
\end{equation}
\begin{equation}
    \alpha_j^{(k)} = \frac{\exp(e_j^{(k)})}{\sum_{n=1}^N \exp(e_n^{(k)})},
\end{equation}
where $v_a$ is the attention vector and $W_a$ is the weight in the attention layer. The context representation of $x^{(k)}$ is then the weighted sum of $h^{(k)}$: 
\begin{equation}
    c^{(k)} = \sum_{j=1}^N \alpha_j^{(k)} h^{(k)}.
\end{equation}

After obtaining $c^{(pre)}$ and $c^{(pos)}$, we concatenate them with the target entity embedding $x^{(r)}$, and pass it through a feed forward network to obtain the final representation:
\begin{equation}
    R = \mbox{ReLU}(W_f [c^{(pre)}, x^{(r)}, c^{(pos)}]),
\end{equation}
where $W_f$ is the weights in the feedforward layer. $R$ is also used as the input of the probing classifiers (section~\ref{sec:prob}).
$R$ is then fed 
for making the final prediction:
\begin{equation}
    P(f|x^{(pre)}, x^{(r)}, x^{(pos)}) = \mbox{Softmax}(W_c R),
\end{equation}
where $W_c$ is the weight in the output layer.

\paragraph{\texttt{c-RNN}.} In addition to \texttt{ConATT}, we also try a simpler yet effective structure, which uses only a single BiGRU. We name the framework it follows as the centred recurrent neural networks (henceforth \texttt{c-RNN}). Specifically, instead of using two separate BiGRUs to encode pre- and pos-contexts, we first concatenate $x^{(pre)}$, $x^{(r)}$, and $x^{(pos)}$, and then encode them together: 
\begin{equation}
    h = \mbox{BiGRU}([x^{(pre)}, x^{(r)}, x^{(pos)}]).
\end{equation}
Suppose that the target entity is in position $i$ of the concatenated sequence, we extract the $i$-th representation from $h_i$ for obtaining $R = \mbox{ReLU}(W_f h_i)$.
After obtaining $R$, the rest of the procedure is the same as \texttt{ConATT}. 

\paragraph{Pre-training.}
As a secondary objective of this study, we want to see whether RFS can benefit from pre-trained word embeddings and language models, whose effectiveness has not yet been explored in REG\footnote{Previously, only~\citet{cao-cheung-2019-referring} used pre-trained embeddings, but no ablation study was done.}. For both \texttt{c-RNN} and \texttt{ConATT}, we try the GloVe embeddings~\citep{pennington-etal-2014-glove} to see how pre-trained word embeddings contribute to the choice of RF. For \texttt{c-RNN}, we try to stake it on the \texttt{BERT}~\citep{devlin-etal-2019-bert} model. In order to let \texttt{BERT} better encode the delexicalised entity labels, we first re-train \texttt{BERT} as a masked language model on the training data of \textsc{w}eb\textsc{nlg}. We then freeze the parameters of \texttt{BERT} and use the model 
to encode the input, which is then fed into \texttt{c-RNN}\footnote{We also explored other ways of using \texttt{BERT}, such as using only \texttt{BERT} plus a feed forward layer to obtain $h$, or not freezing parameters of \texttt{BERT} while training. The resulting models had low performance in all cases.}. 

\paragraph{Machine Learning (ML) based Model}
We used \texttt{XGBoost} \citep{xgboost2016} from the family of Gradient Boosting Decision Trees to train RFS classifiers. 5-fold-cross-validation was used to train the models.
The classifiers were first trained on a wide range of features obtained from the \textsc{w}eb\textsc{nlg} corpus (16 features).  After running a variable importance analysis, we selected a subset of features for the final models. The detailed list of features are presented in Appendix A. 

\begin{figure*}
    \centering
    \includegraphics[scale=0.6]{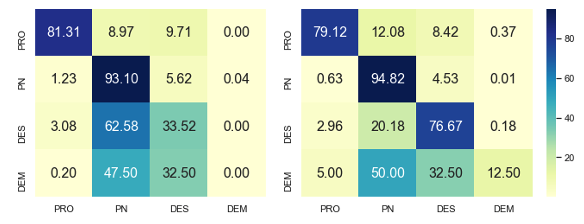}
    \caption{Confusion Matrices for 4-way classification results of \texttt{XGBoost} (left) and \texttt{c-RNN+GloVe} (right), where PRO, PN, DES, and DEM are pronoun, proper name, description and demonstrative respectively.}
    \label{fig:cm}
\end{figure*}

\subsection{Evaluation}

\paragraph{Implementation Details.} 

We tuned hyper-parameters of each of our models on the development set and chose the setting with the best macro F1 score. 
For the \texttt{BERT} model, we used the cased \textsc{bert-base}\footnote{\url{huggingface.co/bert-base-cased}} and added all entity labels into the vocabulary to avoid tokenisation.
When re-training \texttt{BERT} on \textsc{w}eb\textsc{nlg}, we set the masking probability to 0.15 and trained it for 25 epochs.

For the \texttt{XGBoost} models, we set the learning rate to 0.05, the minimum split loss to 0.01, the maximum depth of a tree to 5, and the sub-sample ratio of the training instances to 0.5.

We report the macro averaged precision, recall, and F1 on the test set. We run each model for 5 times, and report the averaged performance.
As for the dataset, we use the v1.5 of \textsc{w}eb\textsc{nlg}~\citep{castro-ferreira-etal-2019-neural} and use only seen entities. 

\paragraph{Results.}

Table~\ref{tab:cls_result} shows the results of different classification tasks. Generally, all neural variants outperform the machine learning baseline. 
The performance difference is small in the case of binary classification, while it is much bigger for 3- and 4-way classifications. 
This is because the 2-way classification (i.e., pronominalisation) has clearly less complexity than the other two alternatives, and, thus, the feature set used by the baseline results in almost similar outcomes to neural models.

Comparing neural variants to each other, the results show that the simpler \texttt{c-RNN} wins over \texttt{ConATT} in 4-way classification, and has on par performance with \texttt{ConATT} for 3- and 2-way classifications. 
One possible explanation is that \texttt{ConATT} first breaks down the input into three pieces (i.e., the target entity as well as pre- and pos-context), encodes them separately, and merges the encoded representations back before being sent to make predictions. 
This ``divide and merge'' procedure might hinder the model from learning some useful information. 

Regarding the effectiveness of incorporating pre-trained models, \texttt{GloVe} embeddings have positive impact on \texttt{c-RNN} only in case of 4-way predictions, and have no contribution to 2- and 3-way classifications. Moreover, it has negative effect on \texttt{ConATT}: the performance diminishes when \texttt{GloVe} is used.
It is surprising to see that in case of \texttt{c-RNN}, \texttt{BERT} has negative effect on 4- and 3-way predictions (the F1 score reduced from 64.86 and 83.63 to 62.15 and 82.15 respectively). 
For pronominalisation, \texttt{BERT} slightly boosts the performance (from 89.09 to 89.42), but this boost is not as much as BERT's boosting effect on other NLP tasks. 
This is probably because although \texttt{BERT} was re-trained on \textsc{w}eb\textsc{nlg} delexicalised sentences, the entity labels still function as noise for \texttt{BERT}.

To obtain insights into the behaviours of the deep learning and classic ML-based models for RFS, we depict the confusion matrices of \texttt{XGBoost} and the best performing neural model \texttt{c-RNN+GloVe} in Figure~\ref{fig:cm} for the 4-way classification. 
The confusion matrices suggest that both models do a good job in selecting pronouns and proper names (that is why the performance difference in the 2-way classification is small), and both perform poorly in choosing demonstratives (probably due to the fact that demonstratives are extremely infrequent in \textsc{w}eb\textsc{nlg}). 
The main difference between the two models is in distinguishing proper names from descriptions. The \texttt{XGBoost} model wrongly predicted the descriptions as proper names in 62.58\% of the cases, while the neural \texttt{c-RNN+GloVe} model did this wrong prediction in 20.18\% of the times. This difference in the performance of the two models might be because the neural models learnt some useful features from the discourse which are not covered in our feature engineering procedure. 
Furthermore, after looking into the \textsc{w}eb\textsc{nlg} dataset, we noticed that various RE cases are annotated incorrectly. For example, \textsc{w}eb\textsc{nlg} annotates ``\emph{United States}'' as a proper name, and ``\emph{the United States}'' as a description. The incorrect annotations might increase the confusion between choosing description and proper name in both \texttt{XGBoost} and \texttt{c-RNN+GloVe}.


\section{Probing RFS models} \label{sec:prob}

We use a logistic regression classifier as our probing classifier. Concretely, for each input, we first use a model discussed in section~\ref{sec:model} to obtain its representation $R$. 
As mentioned in section~\ref{sec:model}, we ran each model five times and reported their averaged scores. For the probing tasks, we use the representations of the models with the best RFS performance on the development set. 

\subsection{Probing Tasks} \label{sec:task}
Following our observations in section~\ref{sec:factor}, we formulate the following probing tasks.

\paragraph{Referential Status.} The referential status of the target entity influences the choice of RF in both linguistic \citep{chafe1976givenness,gundel1993cognitive} and computational studies \citep{castro-ferreira-etal-2016-towards-variation}. 
In this study, we define referential status on two levels: discourse-level and sentence-level. The former (\textbf{DisStat}) has two possible values: (a) discourse-old (i.e., the entity has appeared in the previous discourse) and (b) discourse-new (i.e., the entity has not appeared in the previous discourse). Sentence-level referential status (\textbf{SenStat}) also consists of two values: (a) sentence-new (i.e., the RE is the first mention of the entity in the sentence), and (b) sentence-old (i.e., the RE is not the first mention).

\paragraph{Syntactic Position.} 
Entities in subject position are more likely to be pronominalised than in object position \citep{brennan1995centering, arnold2010speakers}. Therefore, in the syntax probing task (henceforth \textbf{Syn}), we do binary classification: subject or object. 

\begin{table*}[t!]
\small
\centering
\begin{tabular}{llcccccccc}
\toprule
Model & Type & DisStat & SenStat & Syn & DistAnt & IntRef & LocPro & GloPro & MetaPro \\ \midrule
\texttt{Random} & - & \makecell[c]{49.57\\(41.83)} & \makecell[c]{33.11\\(22.87)} & \makecell[c]{49.65\\(48.99)} & \makecell[c]{25.19\\(14.90)} & \makecell[c]{33.30\\(22.92)} & \makecell[c]{50.05\\(49.84)} & \makecell[c]{49.75\\(48.02)} & \makecell[c]{25.24\\(25.20)} \\
\texttt{Majority} & - & \makecell[c]{86.91\\(46.50)} & \makecell[c]{86.91\\(31.00)} & \makecell[c]{61.27\\(37.99)} & \makecell[c]{86.91\\(23.25)} & \makecell[c]{86.91\\(31.00)}  & \makecell[c]{56.28\\(36.01)} & \makecell[c]{68.49\\(40.65)} & \makecell[c]{28.12\\(10.97)}\\ \midrule
\multirow{3}{*}{\texttt{c-RNN}} 
& 4-way & \makecell[c]{85.16\\(84.06)} & \makecell[c]{93.28\\(73.72)} & \makecell[c]{94.16\\(85.34)} & \makecell[c]{92.84\\(53.84)} & \makecell[c]{91.71\\(55.43)} & \makecell[c]{83.37\\(82.92)} & \makecell[c]{70.62\\(56.00)} & \makecell[c]{44.76\\(42.32)} \\
& 3-way & \makecell[c]{84.78\\(83.72)} & \makecell[c]{92.59\\(72.60)} & \makecell[c]{93.50\\(83.60)} & \makecell[c]{92.58\\(54.78)} & \makecell[c]{91.24\\(53.21)} & \makecell[c]{82.17\\(81.67)} & \makecell[c]{70.87\\(56.70)} & \makecell[c]{45.42\\(41.79)} \\
& 2-way & \makecell[c]{88.84\\(88.04)} & \makecell[c]{92.77\\(73.84)} & \makecell[c]{93.49\\(84.00)} & \makecell[c]{92.53\\(54.93)} & \makecell[c]{91.01\\(52.31)} & \makecell[c]{86.08\\(85.69)} & \makecell[c]{71.24\\(59.98)} & \makecell[c]{44.32\\(41.65)} \\ \midrule
\multirow{3}{*}{\makecell[l]{\texttt{c-RNN}\\\texttt{+GloVe}}} 
& 4-way & \makecell[c]{85.84\\(84.85)} & \makecell[c]{93.58\\(74.59)} & \makecell[c]{94.56\\(87.04)} & \makecell[c]{93.30\\(55.67)} & \makecell[c]{92.06\\(55.93)} & \makecell[c]{83.71\\(83.20)} & \makecell[c]{70.55\\(53.53)} & \makecell[c]{44.23\\(41.71)} \\
& 3-way & \makecell[c]{85.09\\(83.89)} & \makecell[c]{91.89\\(67.24)} & \makecell[c]{93.23\\(82.48)} & \makecell[c]{91.72\\(50.94)} & \makecell[c]{90.92\\(51.17)} & \makecell[c]{82.08\\(81.44)} & \makecell[c]{70.20\\(52.49)} & \makecell[c]{45.58\\(42.34)} \\
& 2-way & \makecell[c]{88.88\\(88.02)} & \makecell[c]{92.38\\(71.25)} & \makecell[c]{93.32\\(82.67)} & \makecell[c]{92.25\\(53.67)} & \makecell[c]{90.94\\(51.43)} & \makecell[c]{85.81\\(85.22)} & \makecell[c]{71.78\\(63.17)} & \makecell[c]{44.92\\(41.03)} \\ \midrule
\multirow{3}{*}{\makecell[l]{\texttt{c-RNN}\\\texttt{+BERT}}} 
& 4-way & \makecell[c]{95.85\\(90.64)} & \makecell[c]{94.41\\(78.04)} & \makecell[c]{84.05\\(82.71)} & \makecell[c]{93.60\\(56.91)} & \makecell[c]{92.27\\(54.30)} & \makecell[c]{82.03\\(81.67)} & \makecell[c]{71.04\\(54.24)} & \makecell[c]{45.27\\(43.07)} \\
& 3-way & \makecell[c]{94.00\\(84.80)} & \makecell[c]{92.74\\(72.29)} & \makecell[c]{85.12\\(84.08)} & \makecell[c]{92.57\\(54.21)} & \makecell[c]{91.28\\(53.25)} & \makecell[c]{82.92\\(82.53)} & \makecell[c]{71.69\\(57.31)} & \makecell[c]{43.64\\(42.80)} \\
& 2-way & \makecell[c]{94.59\\(87.28)} & \makecell[c]{92.94\\(69.69)} & \makecell[c]{85.75\\(84.74)} & \makecell[c]{92.50\\(54.19)} & \makecell[c]{92.06\\(54.88)} & \makecell[c]{83.27\\(82.77)} & \makecell[c]{73.80\\(63.07)} & \makecell[c]{41.05\\(40.75)} \\ \midrule
\multirow{3}{*}{\texttt{ConATT}} 
& 4-way & \makecell[c]{94.86\\(87.81)} & \makecell[c]{94.12\\(77.11)} & \makecell[c]{88.64\\(88.00)} & \makecell[c]{93.69\\(57.09)} & \makecell[c]{92.11\\(55.88)} & \makecell[c]{86.93\\(86.34)} & \makecell[c]{72.22\\(60.15)} & \makecell[c]{48.37\\(46.14)} \\
& 3-way & \makecell[c]{93.91\\(84.39)} & \makecell[c]{93.15\\(74.19)} & \makecell[c]{87.43\\(86.66)} & \makecell[c]{92.93\\(55.26)} & \makecell[c]{91.35\\(54.09)} & \makecell[c]{85.32\\(84.56)} & \makecell[c]{72.61\\(60.61)} & \makecell[c]{49.35\\(47.47)} \\
& 2-way & \makecell[c]{93.74\\(84.20)} & \makecell[c]{92.78\\(73.18)} & \makecell[c]{89.01\\(88.44)} & \makecell[c]{92.50\\(53.98)} & \makecell[c]{91.19\\(53.64)} & \makecell[c]{87.05\\(86.75)} & \makecell[c]{70.65\\(56.39)} & \makecell[c]{44.24\\(41.81)} \\ \midrule
\multirow{3}{*}{\makecell[l]{\texttt{ConATT}\\\texttt{+GloVe}}} 
& 4-way & \makecell[c]{94.86\\(87.82)} & \makecell[c]{94.10\\(77.70)} & \makecell[c]{87.98\\(87.24)} & \makecell[c]{93.66\\(57.52)} & \makecell[c]{92.10\\(55.22)} & \makecell[c]{86.06\\(85.69)} & \makecell[c]{71.94\\(58.54)} & \makecell[c]{53.19\\(49.94)} \\
& 3-way & \makecell[c]{93.79\\(84.35)} & \makecell[c]{92.78\\(72.83)} & \makecell[c]{89.54\\(88.91)} & \makecell[c]{92.59\\(54.23)} & \makecell[c]{91.39\\(51.96)} & \makecell[c]{87.09\\(86.80)} & \makecell[c]{71.91\\(59.05)} & \makecell[c]{49.27\\(46.36)} \\
& 2-way & \makecell[c]{93.81\\(84.38)} & \makecell[c]{92.86\\(73.21)} & \makecell[c]{87.69\\(86.96)} & \makecell[c]{92.84\\(56.14)} & \makecell[c]{91.50\\(53.33)} & \makecell[c]{85.61\\(85.27)} & \makecell[c]{72.48\\(62.46)} & \makecell[c]{44.47\\(39.63)} \\
\bottomrule
\end{tabular}
\caption{Results of each probing task. Results are reported in the format of A(B), where A is the accuracy and B is the macro F1.}

\label{tab:prob}
\end{table*}

\paragraph{Recency.} Recency has been used as a vital feature in many of the previous REG or RFS systems \citep{greenbacker2009udel,kibrik2016referential}. It measures the distance between the target entity and its closest antecedent. There are various ways of estimating the recency of a target entity given its context. We hereby use two measures: (1) the number of sentences between the target entity and its antecedent (\textbf{DistAnt}), which consists of four possible values: the entity and its antecedent are (a) in the same sentence, (b) one sentence away, (c) more than one sentence away, and (d) the entity is a first mention (to distinguish first mentions from subsequent mentions). 
(2) whether there is an intervening referent between the target and its nearest antecedent
(\textbf{IntRef}) \citep{greenbacker2009udel}. In other words, it checks whether the target and the preceding RE are coreferential.
 This feature has three possible values: (a) the target entity is a first mention, (b) the previous RE refers to the same entity, and (c) the previous RE refers to a different entity. 
 Note that the existence of intervening markables might signal the existence of a competition (if the intervening referent has the same animacy and gender values as the target RE).

\paragraph{Discourse Structure Prominence.} As mentioned in section \ref{sec:background}, the ``organizational" properties of discourse may influence the prominence status of the entities. We introduce three probing tasks capturing different properties of the discourse. 
(1) \emph{Local prominence} (\textbf{LocPro}): The idea of local prominence is coming from Centering Theory \citep{grosz1995centering}. It is a hybrid feature of DisStat and Syn. Concretely, we use the implementation of \citet{henschel2000pronominalization}: an entity is \emph{locally prominent} if it is ``discourse-old" and ``realised as subject". It is a binary feature with two possible values: (a) locally prominent, and (b) not locally prominent. 
(2) \emph{Global prominence} (\textbf{GloPro}): This feature is based on the notion of global salience in \citet{siddharthan2011information}, asking whether the entity is a minor or major referent in the text. According to them, ``the frequency features are likely to give a good indication of the global salience of a referent in the document" (p. 820). We define a binary feature in which the most frequent entity in a text is marked as globally prominent. 
(3) \emph{Meta-prominence} (\textbf{MetaPro}): In line with global prominence, we also want to explore to what extent prominence beyond a single text (e.g. on a text collection level) may impact the way people refer. In the context of the current circumstances, the sentence ``I received \emph{my vaccine} today" is unambiguous, and the RE \emph{my vaccine} needs no extra modification (e.g. my \textsc{covid-19} vaccine); however, a couple of years from now, a richer RE may be needed to refer to the vaccine. The idea behind this exploratory feature is that people might use less semantic content to refer to the referents which are well known outside of the text. 
Based on the number of mentions of a target entity in the whole \textsc{w}eb\textsc{nlg}, four possible values, each of which representing an interval, are assigned to each RE: (a) $[0, 50)$, (b) $[50, 150)$, (c) $[150, 290)$, and (d) $[290, \infty)$. For example, the category $[0, 50)$ 
contains those entities that occur fewer than 50 times in the corpus.

\subsection{Importance Analysis} \label{sec:importance}

\begin{figure}[t!]
\centering
  \includegraphics[scale=0.30]{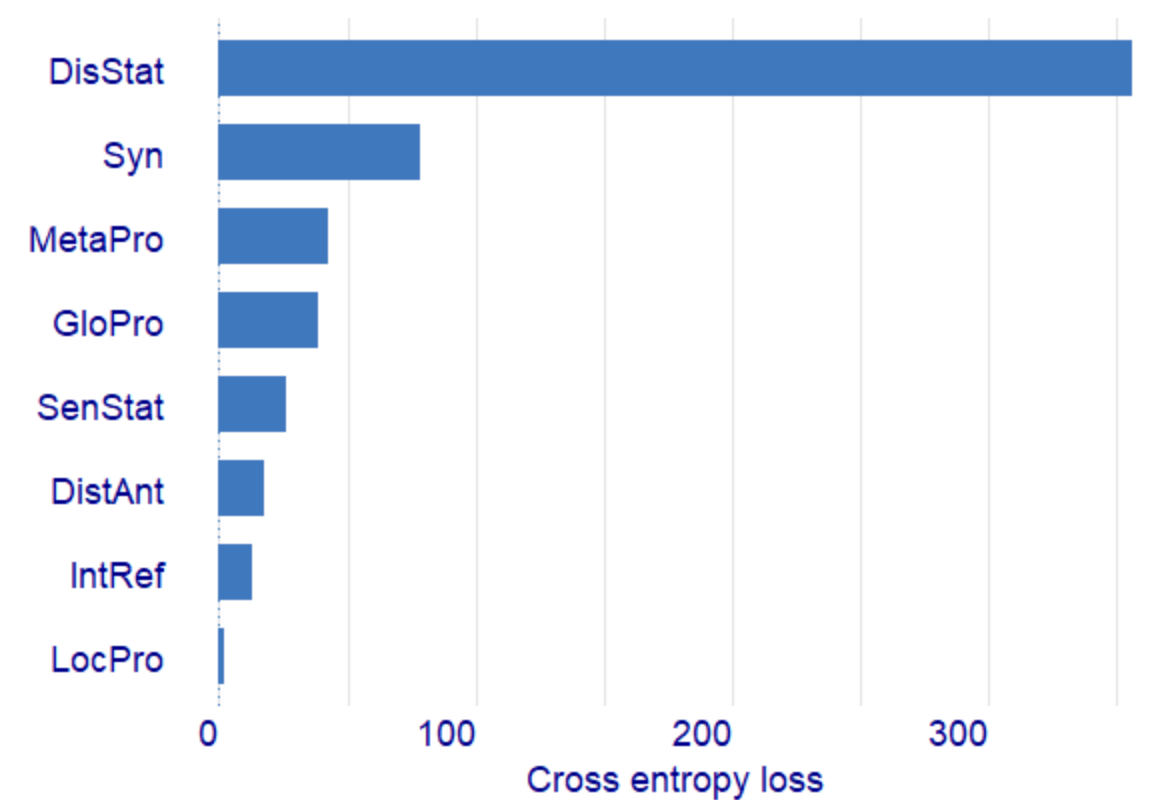}
  \caption{Feature importance of \texttt{XGBoost} classifiers for 4-way predictions. Higher loss shows greater importance of a feature. Results for 2-way and 3-way classification can be found in the Appendix B.}
  \label{fig:ablation}
\end{figure}

We conducted a feature importance analysis to find out which features used in the probing tasks had the highest contributions to the feature-based ML models. This analysis functions as a sanity check to find out whether the representations have learnt the features contributing the most to the RFS task.

To assess the importance of the features used in the probing tasks, we train \texttt{XGBoost} models, only using features from section~\ref{sec:task}, and calculate the model-agnostic permutation-based variable importance of each model \citep{biecek2021explanatory}. Concretely, we measure the extent to which the performance changes if we remove one of the features. 
Figure~\ref{fig:ablation} depicts the performance change for each feature. 
According to the figure, DisStat and Syn contribute the most. LocPro is the least important feature because it is a hybrid combination of DisStat and Syn. Removing it while keeping DisStat and Syn will not hurt the performance of the model a lot. Considering that DisStat and Syn are both highly vital features, LocPro is much more important than what the experiment suggests. In addition to DisStat and Syn probing tasks, we also expect high performance for the LocPro task.

\subsection{Probing Results}
We mentioned earlier that we conduct probing tasks to find out whether the RFS models' latent representations encode the features mentioned in section \ref{sec:task}. High performance in probing tasks would indicate that the features are encoded in the latent representations of the models.

We evaluate probing tasks using the accuracy and macro-averaged F1 scores. Each probing classifier was trained 5 times. We report the averaged value. Additionally, we use 2 baselines: (1) \texttt{random}: it randomly assigns a label to each input; and (2) \texttt{majority}: it assigns the most frequent label in the given probing task to the inputs.

\paragraph{Results of Each Probing Task.} 

Compared to the \texttt{random} baseline, all neural models have achieved higher performance on all tasks. (1) Referential status and syntactic position: all models exhibit consistently high performance on DisStat, SenStat, and Syn. 
This shows that, at least for the \textsc{w}eb\textsc{nlg} corpus, all neural models can learn information of referential status and syntactic position;
(2) Recency (i.e., DistAnt and IntRef): all models perform worse compared to the referential status and syntax probes. 
Although they do not have bad accuracy scores, their F1 scores are lower than that of DisStat, SenStat, and Syn, and are closer to the baselines. 
This finding is consistent with the results of section~\ref{sec:importance}, where DistAnt and IntRef were found to be less important (comparing to DisStat and Syn). One possible explanation is that, in the \textsc{w}eb\textsc{nlg} corpus,  67\% of the documents contain only one sentence, making recency-related features play less role.
As another possible explanation, in line with the previous probing works on coreference and bridging anaphora~\citep{sorodoc-etal-2020-probing,pandit-hou-2021-probing}, models have more difficulty capturing long-distance properties; 
(3) Discourse structure prominence: since LocPro is a hybrid of DisStat and Syn, all models handled it to a large degree. Meanwhile, neural models appear to handle GloPro and MetaPro worse than other features since the performance of their corresponding probing tasks is closer to the baselines\footnote{Note that, for MetaPro, the \texttt{Majority} has low F1 score because the distribution of the values of MatePro is balanced.}. These results are in contrast with the importance analysis results, which suggested that both GloPro and MetaPro are important features (ranking 3 and 4 in Figure~\ref{fig:ablation}). 
Learning GloPro and MetaPro requires a model to have an overall understanding of the whole input document or the whole corpus, which the neural models might not be able to acquire.

\paragraph{Comparing \texttt{c-RNN} and \texttt{ConATT}.}

In section~\ref{sec:model}, we concluded that the \texttt{c-RNN} model works better than \texttt{ConATT} on 4-way RF classification. 
Nevertheless, when probing, we observed that \texttt{ConATT} does a better job in many tasks, including DisStat, LocPro, GloPro, and MetaPro. 
To understand why, we looked into the \textsc{w}eb\textsc{nlg} dataset and found that 86.91\% of the REs in \textsc{w}eb\textsc{nlg} are first mentions, and 21\% of the documents talk about the entity ``\emph{United\_States}''. 
This suggests that REs in \textsc{w}eb\textsc{nlg} are not representative of the realistic use of REs. 
Therefore, although \texttt{ConATT} learns more contextual features, it still has a lower performance.
\texttt{ConATT}'s better learning of referential status (i.e., DisStat) is probably a benefit of using self-attention, which helps the model capture longer dependencies than RNNs.


\paragraph{The Effect of Pre-training.}
As mentioned earlier, the secondary objective of this study is to find out whether RFS can benefit from pre-trained word embeddings and language models.
The effect of incorporating the \texttt{GloVe} embeddings is not significant to \texttt{c-RNN} and \texttt{ConATT}. 
The major contribution of \texttt{BERT} is helping with learning DisStat (which is, again, probably a result of using self-attention). 
Akin to the above discussion, since the majority of the entities in \textsc{w}eb\textsc{nlg} are first mentions, the increased accuracy boost in the DisStat task is not enough to boost the overall performance of RFS.

\paragraph{Comparing Different RF Classifications.}
It also appears that models learn different information using different label sets (classes). For example, 2-way classification (i.e., pronominalisation) helps \texttt{c-RNN} learn more about referential status. But in case of models with attention mechanism (i.e., \texttt{ConATT},  \texttt{ConATT+GloVe} and \texttt{c-RNN+BERT} models), referential status is learnt better in 4-way classification models. 
Also, in case of \texttt{ConATT(+GloVe)}, we observe that more fine-grained classifications help the model learn more about meta prominence (i.e., MetaPro).


\section{Conclusion}

Our aim is to understand whether neural models capture the features associated with the task of RFS. To this end, we defined 8 probing tasks in which we focused on referential status, syntactic position, recency, and discourse structure. The probing results suggest that the probe classifiers always performed better than the \texttt{random} and the \texttt{majority} baselines. The performance was consistently good in the tasks associated with referential status, syntax and local prominence. 

It is worth noting that probing has its own shortcomings. For instance, on the one hand, low probing performance does not always mean the feature is not encoded, but could also mean that such a feature does not matter to RFS.
To mitigate this issue, we conducted a complementary ML-based variable importance analysis; in this analysis, discourse status and syntactic position came out as the factors with the highest contributions. These features were also predicted very well in the probing tasks. However, these results should still be taken with a pinch of salt: the variable importance has been conducted on the ML model and not on the neural models. We cannot be certain that the same features contribute to all the models similarly: a feature might be quite important in the machine learning model, but not as important in the neural models. 
On the other hand, some researches have questioned the validity of probing methods. They found out that it is difficult to distinguish between ``learning the probing task'’ and ``extracting the encoded linguistic information''~\citep{hewitt-liang-2019-designing,kunz-kuhlmann-2020-classifier} for a probing classifier. This suggests that higher performance of a probing classifier does not necessarily mean more linguistic information has been encoded. This prevents us from directly quantifying \emph{how well} the linguistic information has been learnt using the performance of probing classifiers and requires us to make conclusions more carefully.

From our probing efforts, we conclude that: 
(1) All neural models have learnt some information
about the features associated with the probing tasks, but how well they have learnt this information is yet to be assessed; 
(2) The \textsc{w}eb\textsc{nlg} corpus, which has often been used for the study of discourse REG, is not ideally suitable for studying discourse-related aspects of RFS, because the texts are too short and the majority of REs are first mentions.
This leads to bias in the evaluation of RFS and REG algorithms; 
(3) When it comes to the question of how well a RFS feature can be learnt, it matters what neural architecture and label set are used, and whether the model is pre-trained or not.
Using an attention mechanism and more fine-grained label sets help a model learn more information;
(4) All models perform poorly in terms of learning those features, such as GloPro and MetaPro, that do not derive from the text itself but from the wider context in which it is written and read.
We believe that future models should take these lessons into consideration. 

In future, we plan to extend the current study from three angles. First, we plan to conduct experiments on different corpora. The \textsc{w}eb\textsc{nlg} corpus used in this study consists predominantly of extremely short documents with an average length of only 1.4 sentences/document; consequently the majority of REs are first mentions. 
We hope to find a more representative distribution of uses of referring expressions in other corpora such as OntoNotes~\citep{hovy-etal-2006-ontonotes}, which contain longer texts. 
%
Secondly, we plan to conduct experiments on other languages than English, in particular ones that favour zero pronouns (e.g., Chinese~\citep{chen-etal-2018-modelling}), because these pose new challenges for the task of RFS. Thirdly, we plan to 
design new probing tasks on the basis of other factors that could influence RFS, such as animacy, competition and positional attributes (see \citet{same-van-deemter-2020-linguistic} for an overview).


\section*{Acknowledgements}
We thank the anonymous reviewers for their helpful comments.
Guanyi Chen is supported by China Scholarship Council (No.201907720022). Fahime Same is supported by the German Research Foundation (DFG)– Project-ID 281511265 – SFB 1252 “Prominence in Language”.

\bibliographystyle{acl_natbib}
\bibliography{anthology,acl2021}

\appendix
\newpage
\section{Further details on the \texttt{XGBoost} models}
As mentioned earlier, for the RFS task, we firstly created the \texttt{XGBoost} models using a wide selection of features. Afterwards, we ran a variable importance analysis on the models, and chose a smaller subset of features for each classifier. The selected features are presented in Table \ref{tab:feat}. 

\begin{table*}[htbp]
\small
\centering
\begin{tabular}{lllll}
\toprule
Feature & Definition & 2-way & 3-way & 4-way \\ \midrule
\texttt{Syn} & Description is provided in the main text. & \cmark & \cmark & \cmark  \\
\texttt{Entity} & Values: Person, Organisation, Location, Number, Other & \cmark & \cmark & \cmark  \\
\texttt{Gender} & Values: male/female/other & \cmark & \cmark & \cmark  \\
\texttt{DisStat} & Description is provided in the main text.& \cmark & \cmark & \cmark  \\
\texttt{SenStat} & Description is provided in the main text. & - & \cmark & \cmark  \\
\texttt{DistAnt\_S} & Description is provided in the main text (DistAnt). & \cmark & \cmark & \cmark  \\
\texttt{DistAnt\_W} & Distance in number of words (5 quantiles) & \cmark & - & \cmark  \\
\texttt{Sent\_1} & Does RE appear in the first sentence? & \cmark & \cmark & \cmark  \\
\texttt{MetaPro} & Description is provided in the main text. & \cmark & \cmark & \cmark  \\
\texttt{GloPro} & Description is provided in the main text. & \cmark & \cmark & \cmark  \\
\bottomrule
\end{tabular}
\caption{Features used in the \texttt{XGBoost} models.}
\label{tab:feat}
\end{table*}

\section{Further results of importance analysis}
\begin{figure*}
    \centering
    \includegraphics[scale=0.5]{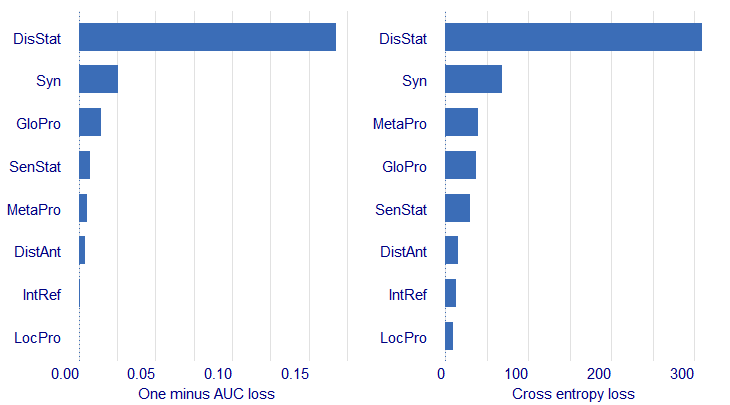}
    \caption{Feature Importance of the \texttt{XGBoost} 2-way (left figure) and 3-way (right figure) predictions. }
    \label{fig:varimp_2_3}
\end{figure*}

Figure \ref{fig:varimp_2_3} depicts the variable importance results for the 2-way and 3-way classification tasks. As mentioned in the paper, there is a high degree of agreement between the ordering of the variables in the 3 models. 

\begin{figure*}
    \centering
    \includegraphics[scale=0.4]{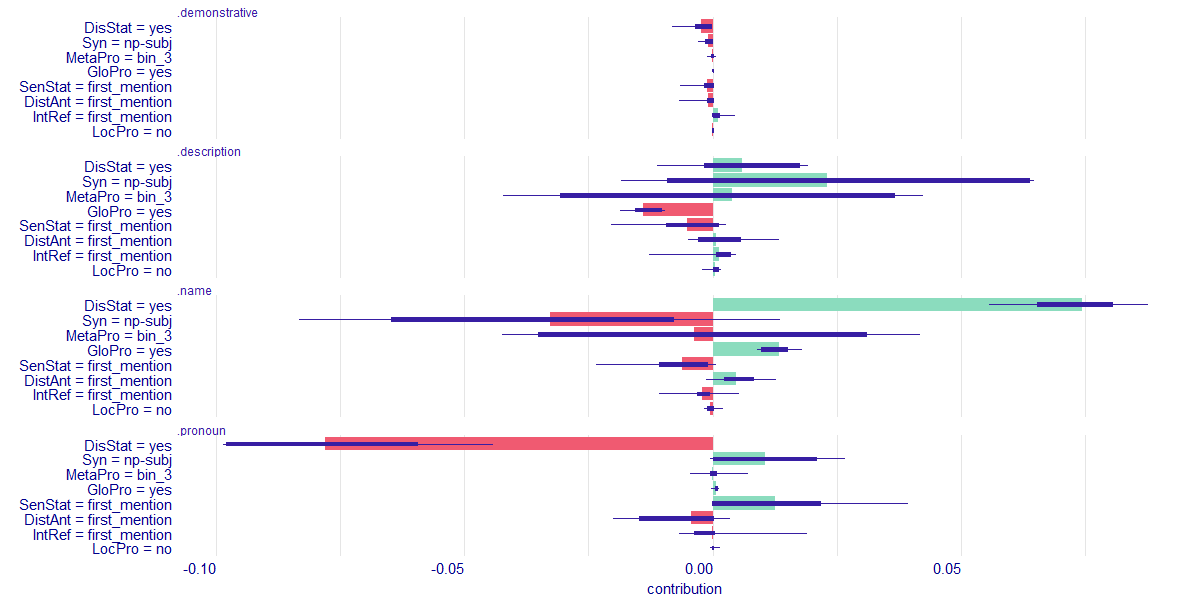}
    \caption{Shapley values with box plots for 100 random orderings of explanatory variables in the \texttt{XGBoost} 4-class model.}
    \label{fig:shap_4}
\end{figure*}

To get a better idea about the contribution of each variable to the decisions made by the models, Figure \ref{fig:shap_4} demonstrates the shapley values for 100 random orderings of explanatory variables in the 4-way classification model. The figure clearly shows that the model has failed to learn the \texttt{demonstrative} class. For other decisions, the model majorly uses 2 features, namely \texttt{DisStat} (referential status) and \texttt{Syn} (syntactic role).

\end{document}